%% file: main.tex
\documentclass[10pt]{article}

\pdfoutput=1
\input{preamble.tex}

\title{Case-based off-policy policy evaluation\\using prototype learning}
\author{Anton Matsson and Fredrik D. Johansson}
\affil{Chalmers University of Technology}
\date{}

\begin{document}

\maketitle

%%%%%%%%%%%%%%%%%%%%%%%%%%%%%%
%%%%%%%%%% Abstract %%%%%%%%%%
%%%%%%%%%%%%%%%%%%%%%%%%%%%%%%

\begin{abstract}
Importance sampling (IS) is often used to perform off-policy policy evaluation but is prone to several issues---especially when the behavior policy is unknown and must be estimated from data. Significant differences between the target and behavior policies can result in uncertain value estimates due to, for example, high variance and non-evaluated actions. If the behavior policy is estimated using black-box models, it can be hard to diagnose potential problems and to determine for which inputs the policies differ in their suggested actions and resulting values. To address this, we propose estimating the behavior policy for IS using prototype learning. We apply this approach in the evaluation of policies for sepsis treatment, demonstrating how the prototypes give a condensed summary of differences between the target and behavior policies while retaining an accuracy comparable to baseline estimators. We also describe estimated values in terms of the prototypes to better understand which parts of the target policies have the most impact on the estimates. Using a simulator, we study the bias resulting from restricting models to use prototypes.
\end{abstract}

%%%%%%%%%%%%%%%%%%%%%%%%%%%%%%%%%%
%%%%%%%%%% Introduction %%%%%%%%%%
%%%%%%%%%%%%%%%%%%%%%%%%%%%%%%%%%%

\section{Introduction}
\label{sec:intro}

Historical data on decisions and outcomes provide opportunities for evaluating  policies for future decision-making. For example, the prospect of using patient records to evaluate new policies for medication dosing in sepsis management has attracted recent attention~\citep{komorowski2018artificial,gottesman2019guidelines}. An example of off-policy policy evaluation (OPPE), this amounts to estimating the value of a target policy based on data gathered under a different so-called behavior policy; see e.g.,~\citet{thomas2015safe} for an overview.

\begin{figure}[t!]
\centering
\includegraphics[width=0.5\linewidth]{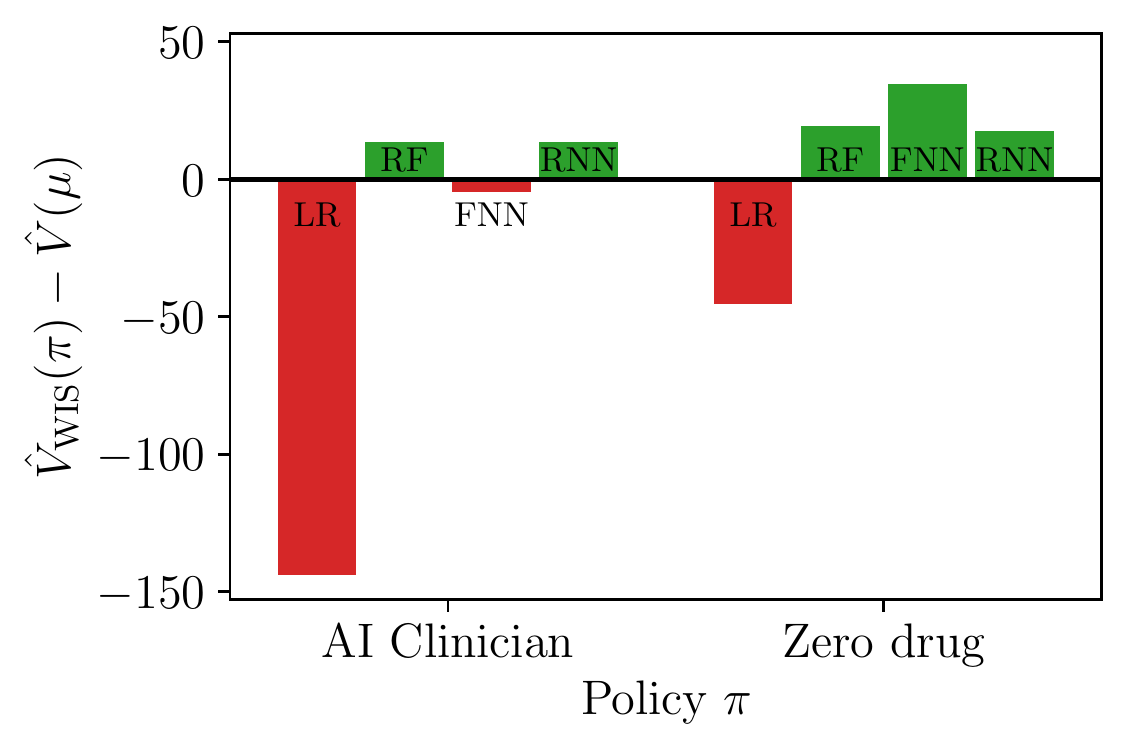}
\caption{Off-policy policy evaluation of two target policies for sepsis management: the so-called AI Clinician~\citep{komorowski2018artificial} and a zero-drug policy. Weighted importance sampling of observations from MIMIC-III~\citep{mimiciii} was used to estimate the value of each policy based on several models of the unknown behavior policy $\mu$, followed by physicians in data: a logistic regression classifier (LR), a random forest classifier (RF), a feedforward neural network (FNN) and a recurrent neural network (RNN). For each estimator, we plot the difference between the value estimates of behavior and target policies. The results suggest that both target policies may be superior to the behavior policy, especially the zero-drug policy. However, never treating patients with sepsis clearly goes against intuition. With this in mind, can we trust the estimates of the policies' values?} 
\label{fig:intro}
\end{figure}

Importance sampling (IS) methods~\citep{precup2000eligibility} perform OPPE for decision-making by weighting observed outcomes by the density ratio of the target policy and the behavior policy. IS methods are often preferred over alternatives, which rely on modeling outcomes or covariate transitions, due to their simplicity and that behavior policies often are controllable or human made. Similarly, the equivalent strategy of inverse-propensity weighting is fundamental to the study of causal effects~\citep{rosenbaum1983central,hirano2003efficient}.

In practice, it is difficult to assess the quality of an IS value estimate; see Figure~\ref{fig:intro} for an example. When the  behavior policy is unknown and must be estimated from data, conditions which guarantee good estimates are hard to meet and rely on untestable assumptions~\citep{rosenbaum2010design,namkoong2020off}. Standard practices of inspecting weights \citep{li2019addressing} or removing outliers \citep{crump2009dealing} give only aggregate or per-sample perspectives on potential issues. These are often insufficient for a domain expert to reason about the result's validity.

In this work, we study OPPE of sequential decision-making policies using importance sampling with an unknown behavior policy. To enable diagnosis of potential issues in cases where the input space is large or sequential, we propose estimating the behavior policy using prototype learning~\citep{li2018deep,ming2019prototypes}. Prototypes are cases from the input data, selected by the learning algorithm, which are readily interpretable by a domain expert and representative of the state-action space. In healthcare applications, prototypes correspond to trajectories of former patients, and prototype-based policies to how physicians use experience from previous patients to treat new ones.

We show that using a prototype model for OPPE has several advantages over models which  output only a point estimate of the behavior policy: a) case-based estimates make it easier for domain experts to assess their validity;  b) prototypes can be used to describe areas of low overlap between behavior and target policies; c) prototypes induce a clustering which may be used to explain differences in value for different policies.

To illustrate our method, we study the management of sepsis, for which several AI-based policies have been proposed \citep{peng2018improving, komorowski2018artificial}. Examining the target policy of \citet{komorowski2018artificial}, we show that a prototype-based estimate of the behavior policy allows us to more easily describe differences between behavior and target policies, spot violations of overlap and explain which patients contribute to differences in estimated values. Further, we use a simulator of sepsis patients to quantitatively estimate the added bias of restricting our model class to use prototypes, as the number of time points increases.

%%%%%%%%%%%%%%%%%%%%%%%%%%%%%%%%%%%%%%%%%%%%%%%%%%
%%%%%%%%%% Off-policy policy evaluation %%%%%%%%%%
%%%%%%%%%%%%%%%%%%%%%%%%%%%%%%%%%%%%%%%%%%%%%%%%%%

\section{Off-policy policy evaluation}

Policy evaluation refers to estimating the \emph{value} $V(\pi)$ of a \emph{target policy} $\pi \in \Pi$. We focus on the sequential case, where a policy is used to select an \emph{action} $A\in\cA=\{1, \ldots, k\}$ after a \emph{history} $H \in \cH$, comprising a sequence of previous actions and \emph{contexts} $X\in\cX$. The history until time $t$ is defined as $H_{t}\coloneqq(X_{0}, A_{0}, X_{1}, A_{1}, \ldots, X_{t})$, with $H_{0}=X_{0}$. A policy $\pi : \cH \rightarrow \Delta_\cA$ is a map from a history to a distribution over $\cA$. In a medical example, a context $X$  contains information about a patient's state, an action $A$ is a medical intervention, and the target policy $\pi$ corresponds to new clinical guidelines.

The value of a policy $\pi$ is defined as the expectation of a \emph{reward} or \emph{outcome} $R\in \bbR$, accumulated after acting according to $\pi$. Here, we study the special case where a single reward is awarded at the end of the sequence, $R = R_T$, but our results generalize to the case where rewards are given after every action. Under the distribution $p_\pi(X_0, A_0, \ldots, X_T, A_T, R) = p_\pi(H_T, A_T, R)$, induced by the policy $\pi$, the value is $V(\pi) \coloneqq \E_\pi[R]$.

Estimating $V(\pi)$ is trivial given a large enough number of samples from $p_\pi$. In off-policy policy evaluation (OPPE), we have access to no such samples, but must estimate $V(\pi)$ using an observational dataset of $m$ samples $\cD = ((h^1_{t_1}, a^1_{t_1}, r^1), \ldots, (h^m_{t_m}, a^m_{t_m}, r^m))$, drawn according to a distribution  $p_{\mu}(H_T,A_T,R)$, controlled by a \emph{behavior policy} $\mu \in \Pi$. In the medical example, the behavior policy represents current clinical practice. In this work, the behavior policy $\mu$ is unknown and an estimate  $\hat{p}_{\mu}(A\mid H)$ is learned from the samples $\cD$.

A common method for OPPE is \emph{importance sampling} (IS).\footnote{Importance sampling estimators are often also referred to as importance weighting estimators.} The (standard) importance sampling estimator uses an estimate $\hp_\mu$ in a weighted average over the samples $\cD$~\citep{hanna2019importance}:
\begin{eqnarray}
    & \displaystyle{
        \hat{V}_\IS(\pi; \hat{\mu}) 
        \coloneqq 
        \frac{1}{m}\sum_{i=1}^{m}w_{i}r^{i},
        \label{eq:VIS}
    }
    \\
    \mbox{with} \;\; 
    & \displaystyle{
        w_{i}
        \coloneqq
        \prod_{t=0}^{t_{i}}
        \frac{p_{\pi}(A_{t}=a_{t}^{i} \mid H_{t}=h_{t}^{i})}
        {\hat{p}_{\mu}(A_{t}=a_{t}^{i} \mid H_{t}=h_{t}^{i})}.
        \label{eq:wr_redef}
    }
\end{eqnarray}
Under appropriate  assumptions (see Section \ref{sec:is_problems} for more details), the estimator $\hat{V}_\IS(\pi;\hat{\mu})$ is an unbiased estimator of $V(\pi)$ but suffers from high variance when $p_\mu$ and $p_\pi$ differ significantly. The weighted importance sampling estimator~\citep{rubinstein2016simulation}, 
$\hat{V}_\WIS(\pi; \hat{\mu}) \coloneqq \frac{1}{\sum_{i=1}^{m}w_{i}}\sum_{i=1}^{m}w_{i}r^{i}$,
introduces bias, but often has less variance. Under the Markov assumption, i.e., that context (or ''state'') transitions, actions and rewards depend only on the most recent context-action pair, the history $H_{t}$ in \eqref{eq:wr_redef} can be replaced with the context $X_{t}$. We leave out the subscript $t$ where clear.

%%%%%%%%%% Trusting IS %%%%%%%%%%%

\subsection{Can we trust an IS estimate?}
\label{sec:is_problems}

If a value estimate of a policy $\pi_1$ is higher than that of another policy $\pi_2$, why is that so? Is it due to the superiority of $\pi_1$, due to random chance, or due to systematic errors in the IS estimates? If the former is true, in which situations does $\pi_1$ yield better actions?

Sufficient conditions for unbiased estimation include (sequential) ignorability and overlap~\citep{rosenbaum1983central,robins1986new}. As ignorability is untestable, we focus on overlap here. Overlap is satisfied for a history-action pair $(h, a)$ if it being observable under $\pi$ implies that it is observable under $\mu$, that is, for all $t$,
\begin{equation*}
p_{\pi}(A_t=a \mid H_t=h) > 0 \Rightarrow p_{\mu}(A_t=a \mid H_t=h) > 0.
\end{equation*}
A fundamental problem is that the extent of overlap is unknown if $\mu$ is unknown. As a result, assessing the quality of an estimate $\hat{V}_\IS(\pi; \hat{\mu})$ \emph{inherently relies on evaluation by a domain expert}. It is critical that the OPPE method is transparent enough to enable such evaluation.

When examining IS estimates, it is common to inspect the importance weights $\{w_i\}_{i=1}^m$ or estimated action propensities $\hat{p}_\mu(A_t \mid H_t)$ directly~\citep{li2019addressing}. This enables the analyst to spot outliers with extremely large weights, or to compute the effective sample size (ESS)~\citep[Chapter 9]{gottesman2018evaluating,mcbook}, giving a per-sample and an average view of potential issues with variance and the potential for removing samples with excessive weights~\citep{crump2009dealing,sturmer2010treatment}. However, these practices leave several questions unanswered:

\begin{enumerate}[A.]
\item \textbf{Which observations contribute to the IS estimate?} What \emph{signifies} the trajectories and history-action pairs that are assigned high importance weights, $\{ i : w_i \gg 1\}$?
\item \textbf{In which situations is overlap violated?} When does $\pi$ suggest actions $a$ that would never be observed under $\mu$? Can we describe the set $\{(h,a) : p_\pi(a \mid h)>0$,  $\hp_\mu(a \mid h) \approx 0\}$? This is not detectable in $\{w_i\}$ since observing a sample from this set has probability $\approx 0$.
\item \textbf{If $\hat{V}(\pi) > \hat{V}(\mu)$, what gives $\pi$ the edge?} In \emph{which} situations does acting according to $\pi$ result in higher rewards than acting according to $\mu$? 
\end{enumerate}

Inspecting weights and propensities in aggregate or on a per-sample basis is insufficient to answer questions A--C as they all concern \emph{patterns} in policy decisions, weights and rewards. Next, we show how case-based estimates of the behavior policy $\mu$ can help identify such patterns for interpretation by a domain expert.

%%%%%%%%%%%%%%%%%%%%%%%%%%%%%%%%%%%%%%%%%%
%%%%%%%%%% OPPE with prototypes %%%%%%%%%%
%%%%%%%%%%%%%%%%%%%%%%%%%%%%%%%%%%%%%%%%%%

\section{Off-policy policy evaluation with prototypes}
\label{sec:is_proto}

We propose performing off-policy policy evaluation using \emph{prototype learning}~\citep{li2018deep,ming2019prototypes}. The idea is to express the probability  $p_{\mu}(A \mid H)$ by comparing the history $H$ to a relatively small set of prototype cases---histories---from the training data. In a clinical setting, such a policy corresponds to physicians choosing treatment for a new patient based on their prior experience in treating similar patients. For a domain expert, trained in interpreting such cases, a prototype-based estimate is transparent as long as the number of prototypes is small enough. By examining how importance weights, policy overlap and value estimates vary for different prototypes, we obtain answers to questions A--C raised above.

%%%%%%%%%% Prototype learning %%%%%%%%%%

\subsection{Modeling the behavior policy with prototypes}
\label{sec:protomodel}

Let $\tH = [\th^{1}, \ldots, \th^{n}]^{\intercal}$ be a list of $n$ prototype histories\footnote{From now, we refer to these as ''prototypes''.}, each element of which is a \emph{subsequence} of an observed history, $\th^j = h_t^i$ for $h^i \in \cD$ and $t \leq t_i$. Prototypes are used to approximate the behavior policy $p_\mu(A_{t} \mid H_{t}=h_{t})$ based on the similarity between observations $h_t$ and prototypes in a learned representation. The prototypes are themselves selected by the learning algorithm. 

To learn $\tH$, we follow \citet{li2018deep, ming2019prototypes} by first learning a set of latent prototypes as free parameters $\tZ = \left[\tz_{1}, \ldots, \tz_{n}\right]^{\intercal}$ in an encoding space $\cZ$. These are later projected onto encoded observations from $\cD$ to select prototypes. Given an encoder $e : \cH \rightarrow \cZ$, for an arbitrary history $h_{t}$, let
\begin{equation*}
S(\tZ, e(h_{t})) = [s(\tz_{1}, e(h_{t})), \ldots, s(\tz_{n}, e(h_{t}))]^{\intercal}
\end{equation*}
be the \emph{similarity vector} for the encoding of $h_{t}$ comparing $e(h_{t})$ to $\tZ$ using a fixed function $s : \cZ \times \cZ \rightarrow \bbR_{+}$. 
With $B \in \bbR^{k \times n}$, we estimate the behavior policy $\mu$ through  logistic regression in the space induced by $S$, 
\begin{equation}
    \hp_\mu(A_{t} \mid H_{t}=h_t) = \sigma(BS(\tZ, e(h_t)) + c),
    \label{eq:softmax}
\end{equation}
where $\sigma$ denotes the softmax function over rows and $c \in \mathbb{R}^k$ is a bias term. Column $j$ of $B$ represents the probability coefficients associated with $\th^j$. If the coefficient $B_{ij}$ is positive, similarity between $h_{t}$ and $\th^j$ makes action $i$ more probable for $h_{t}$; a negative coefficient makes action $i$ less probable. 

The model parameters $\Theta = (e, B, c, \tH)$, comprising the parameters of the encoder $e$, coefficients $B$, $c$ and the set of prototypes $\tH$, are all unknown and must be learned from data. Following \citet{ming2019prototypes}, we do this by minimizing the regularized negative log-likelihood of the observed data $\cD$ under the model in \eqref{eq:softmax}:
\begin{equation}
    J(\Theta) = \mathrm{NLL}(\cD; \Theta) + \lambda_{d}R_{d}(\Theta) + \lambda_{c}R_{c}(\Theta) + \lambda_{e}R_{e}(\Theta).
    \label{eq:objective}
\end{equation}
Regularization terms $R_{d}(\Theta)$, $R_{c}(\Theta)$ and $R_{e}(\Theta)$ encourage \emph{diversity}, \emph{clustering} and \emph{evidence}, respectively. All terms are fully defined in Appendix~\ref{app:reg}. $J(\Theta)$ is minimized using stochastic gradient descent. To make sure that prototypes represent real cases, i.e., to select $\tH$, latent prototypes are projected onto encodings of training samples at regular intervals between descent steps,
\begin{equation}
    \th^j \leftarrow \argmin_{h^i_t \in \overline{\cD}} \, s(\tz_{j}, e(h^i_t))
    \;\;\; \mbox{ and } \;\;\; 
    \tz_{j} \leftarrow e(\th^j),
    \label{eq:projection}
\end{equation}
with $\overline{\cD}$ the set of all subsequences of trajectories in $\cD$.

We consider two parameterizations of encoding functions $e$, a feedforward neural network (FNN) for one-time decision-making and Markovian policies, and a recurrent neural network (RNN) for non-Markovian policies. For the fixed similarity function $s$ we use an RBF-kernel with unit bandwidth, 
\begin{equation*}
s(\tz, e(h)) \coloneqq \exp(-\|\tz - e(h)\|_2^2).
\end{equation*}

%%%%%%%%%% Prototype predictions %%%%%%%%%%%

\subsection{Predicting with prototypes}
\label{sec:prediction}

When computing the estimated behavior policy \eqref{eq:softmax} for a history $h$, the similarity vector $S(e(\tH), e(h))$ determines how similar each of the $n$ prototypes are to $h$. The number $n$ is a hyperparameter. The more prototypes are used, the greater the flexibility of the model, but a large $n$ may result in $S$ consisting of multiple elements close to $1$, making predictions more difficult to interpret. For example, if $s(e(\th^j), e(h))\approx 1$ for more than 10 prototypes $j$, it may be difficult to reason about the policy decision after all. 

To mitigate this problem, we use only a limited number of  $q\leq n$ prototypes---so-called prediction prototypes---when making predictions with the trained model. Let $s_{q}(h)$ be the similarity between $e(h)$ and its $q$th most similar latent prototype. For $j=1, \ldots, n$, we truncate the similarity vector according to
\begin{equation*}
    s(\tz_{j}, e(h)) \leftarrow 
	\left\{ 
	\begin{array}{ll}
    	s(\tz_{j}, e(h)) &\text{if }\; s(\tz_{j}, e(h)) \geq s_{q}(h), \\
        0 &\text{otherwise}.
    \end{array}
    \right.
\end{equation*}
We perform this step independently for all contexts $h$. In experiments, we study the resulting trade-off between transparency (small $q$ and $n$) and bias as we vary the number of (prediction) prototypes, ($q$) $n$.

%%%%%%%%%% Prototype values %%%%%%%%%%

\subsection{Inspecting prototype-based estimates}
\label{sec:protovalues}

The histories $\th^j$ of prototypes $j=1, \ldots, n$, their policy coefficients $B_j$, and estimated action probabilities 
\begin{equation}
\hp_j(a) = \hp_\mu(A=a \mid H=\th^j)
\label{eq:prototype_policy}
\end{equation}
give an overview of the behavior policy estimate. It allows for describing context regions where the probability of certain actions is low ({\bf Questions A and B}) and for describing differences between $p_\mu(A\mid H)$ and $p_\pi(A\mid H)$.

As we show below, using prototypes also allows for stratifying the value function $V(\pi)$ to describe which histories contribute to differences in estimated value between two policies $\pi$ and $\mu$. While it is possible to use any soft or hard clustering of the space of histories for this, prototypes have the advantage of being based in cases, and trained to describe groups of subjects who are treated differently under the behavior policy.

We define $J_t$ to be a random variable with values in $\{1, \ldots, n\}$, representing an assignment of a history $H_t$ to a prototype at time $t$. We let $J_t$ be distributed according to a normalization of the similarity $s$,
\begin{equation*}
p(J_t=j \mid H_t=h) = \frac{s(\tilde{z}_j, e(h))}{\sum_{k=1}^n s(\tilde{z}_k, e(h))}.
\end{equation*}
Based on this, we define the value $V_{j,t}(\pi)$ of prototype $j$ at time $t$, obtained under a policy $\pi$, as the expected future reward under $\pi$ given the assignment $J_t = j$,
\begin{equation}
V_{j,t}(\pi) \coloneqq \E_{\pi} [R_T \mid J_t = j ].
\label{eq:prototype_value}
\end{equation}

With $p(J_t=j)$ the marginal probability of being assigned to prototype $j$ at time $t$, by the law of total expectation, $V(\pi) = \sum_{j=1}^n V_{j,t}(\pi) p(J_t=j)$ for any $t$. Each term $j$ in the sum represents the contribution to the overall value $V(\pi)$ from histories which are similar to prototype $j$ at time $t$, effectively stratifying the value by types of situations ({\bf Question C}).

We may express $V_{j,t}(\pi)$ as a weighted expectation under the behavior policy $\mu$, with importance weights $W$,
\begin{equation*}
V_{j,t}(\pi) \coloneqq \E_{\mu} \bigg[\frac{p(J_t=j \mid H_t)}{p_\pi(J_t=j)}  W R_T  \bigg],
\end{equation*}
where $p_\pi(J_t)$ is found by importance-weighted marginalization over $H_t$ (see derivation in Appendix~\ref{app:protvalue}). We use this strategy to estimate $V_{j,t}(\pi)$ from finite samples.

%%%%%%%%%%%%%%%%%%%%%%%%%%%%%%%%%
%%%%%%%%%% Experiments %%%%%%%%%%
%%%%%%%%%%%%%%%%%%%%%%%%%%%%%%%%%

\section{Experiments}

We illustrate our method by further examining the example of sepsis management introduced in Figure~\ref{fig:intro}, first using patient data from the MIMIC-III database~\citep{mimiciii}, and second using a sepsis simulator where true policy values are known. 

In the MIMIC-III case, we focus on evaluating a replication of the target policy of the so-called AI Clinician implemented by \cite{komorowski2018artificial}. The data (e.g., demographics, vital signs and laboratory values) are coded as multidimensional time series with a discrete time step of 4 hours. In short, AI Clinician is learned by clustering the data into 750 states, discretizing combinations of intravenous (IV) fluids $(f)$ and vasopressors $(v)$ into 25 possible actions $(f, v) \in \{0, 1, 2, 3, 4\}^2$, and solving the corresponding Markov decision process using value iteration with rewards $r^{i}=\pm 100$ based on the survival of the patients. The process is repeated 500 times, each time with a new train-test split, and the best performing policy on the test set is taken as the target policy, $\pi$. We return to the sepsis simulator in Section~\ref{sec:simulator}.

%%%%%%%%%% Policy evaluation %%%%%%%%%%%

\subsection{Evaluating policies}

\begin{figure}
\centering
\begin{minipage}{.48\textwidth}
  \centering
  \includegraphics[width=\linewidth]{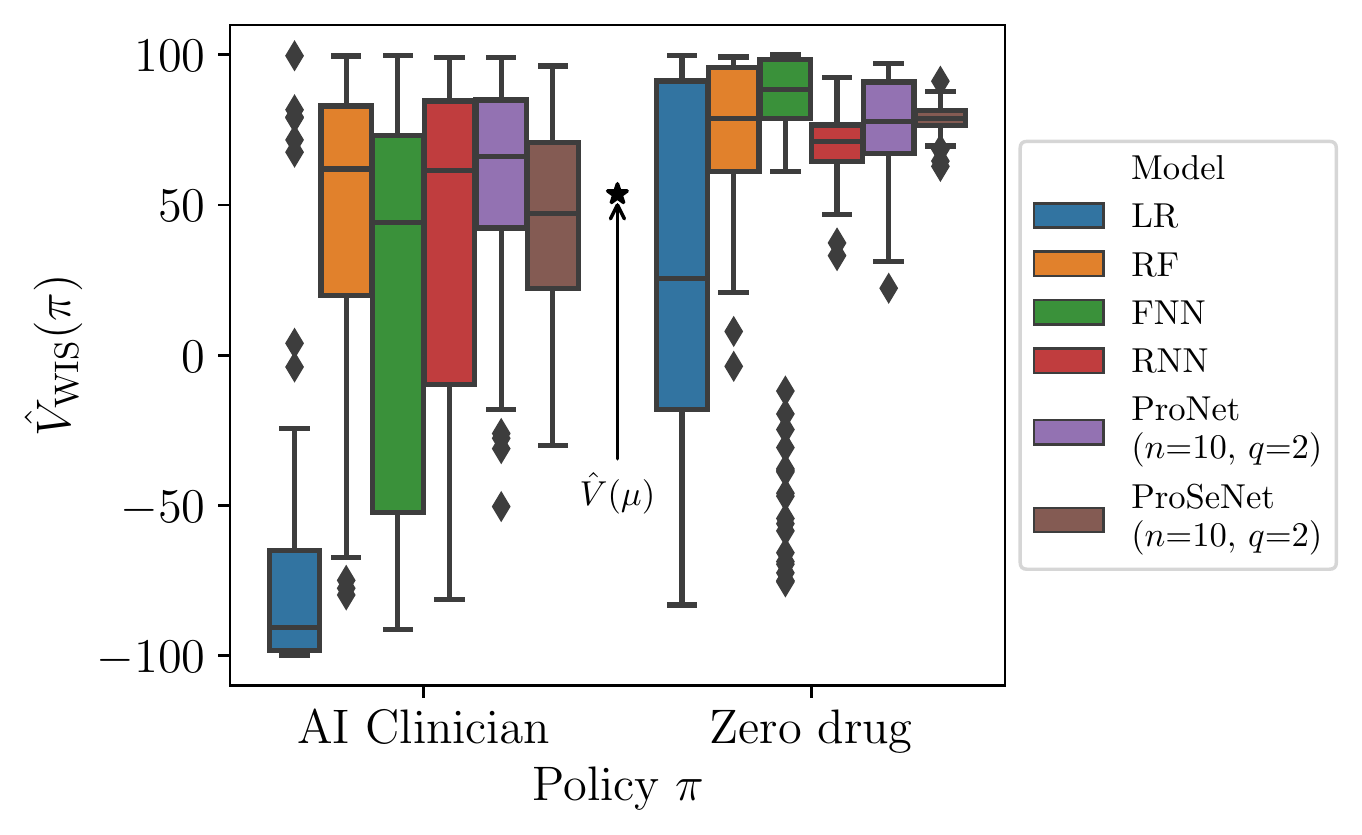}
  \caption{Estimated values of the AI Clinician and a zero-drug policy on the test data. The estimated value of the behavior policy $\mu$ is included as a reference.}
  \label{fig:sepsis:wis}
\end{minipage}
\hfill
\begin{minipage}{.48\textwidth}
  \centering
  \includegraphics[width=\linewidth]{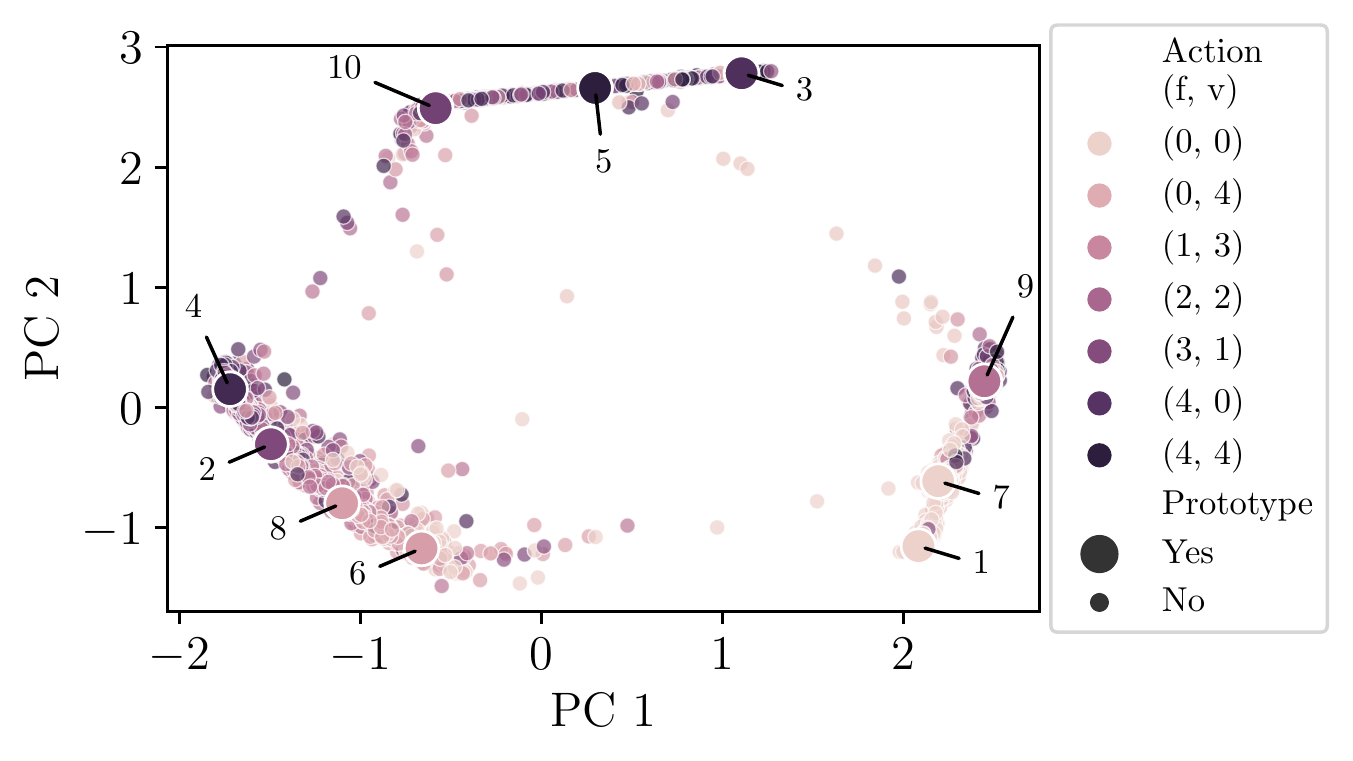}
  \caption{A PCA plot of encoded training samples, colored  w.r.t. the action taken by the physicians. The prototypes are numbered 1--10.}
  \label{fig:sepsis:pca_action}
\end{minipage}
\end{figure}

The results from Figure~\ref{fig:intro} are reproduced with greater detail in Figure \ref{fig:sepsis:wis}. Here, we show 100 repeated WIS estimates of the value of the AI Clinician and the zero-drug policy using different estimators of the behavior policy, $\mu$. The estimated value of $\mu$, $\hat{V}(\mu)$, is included as a reference. As baseline estimators of $p(A\mid H)$, we use a logistic regression classifier (LR), a common choice for propensity estimation, a random forest classifier (RF), a feedforward neural network (FNN) and a recurrent neural network (RNN). We compare these to two prototype models---ProNet and ProSeNet---with an FNN encoder and an RNN encoder, respectively. For all models except RNN and ProSeNet, we make the Markov assumption and model $p(A \mid H)$ using only the last context-action pair of the history.

For both prototype models we use $n=10$ prototypes and $q=2$ prediction prototypes (see Appendix B for further implementation details). As we concluded already in Figure~\ref{sec:intro}, the results seem to indicate that both the AI Clinician and the zero-drug policy are better than the behavior policy followed by physicians. With the exception of LR, all estimators return comparable results. However, leaving all patients untreated is clearly a bad idea, and we should not accept this result without further examination.

Inspecting learned prototypes provides a compact summary of differences between $\mu$ and $\pi$, and how these contribute to the value estimates. To make this precise, we study the prototypes of the ProSeNet model. As an overview of the relationship between prototypes, a PCA plot of encoded training data, including the latent prototypes, numbered 1--10, is shown in Figure \ref{fig:sepsis:pca_action}, where the colors indicate treatment chosen by $\mu$.

\begin{figure}[t!]
\centering
\includegraphics[width=0.5\columnwidth]{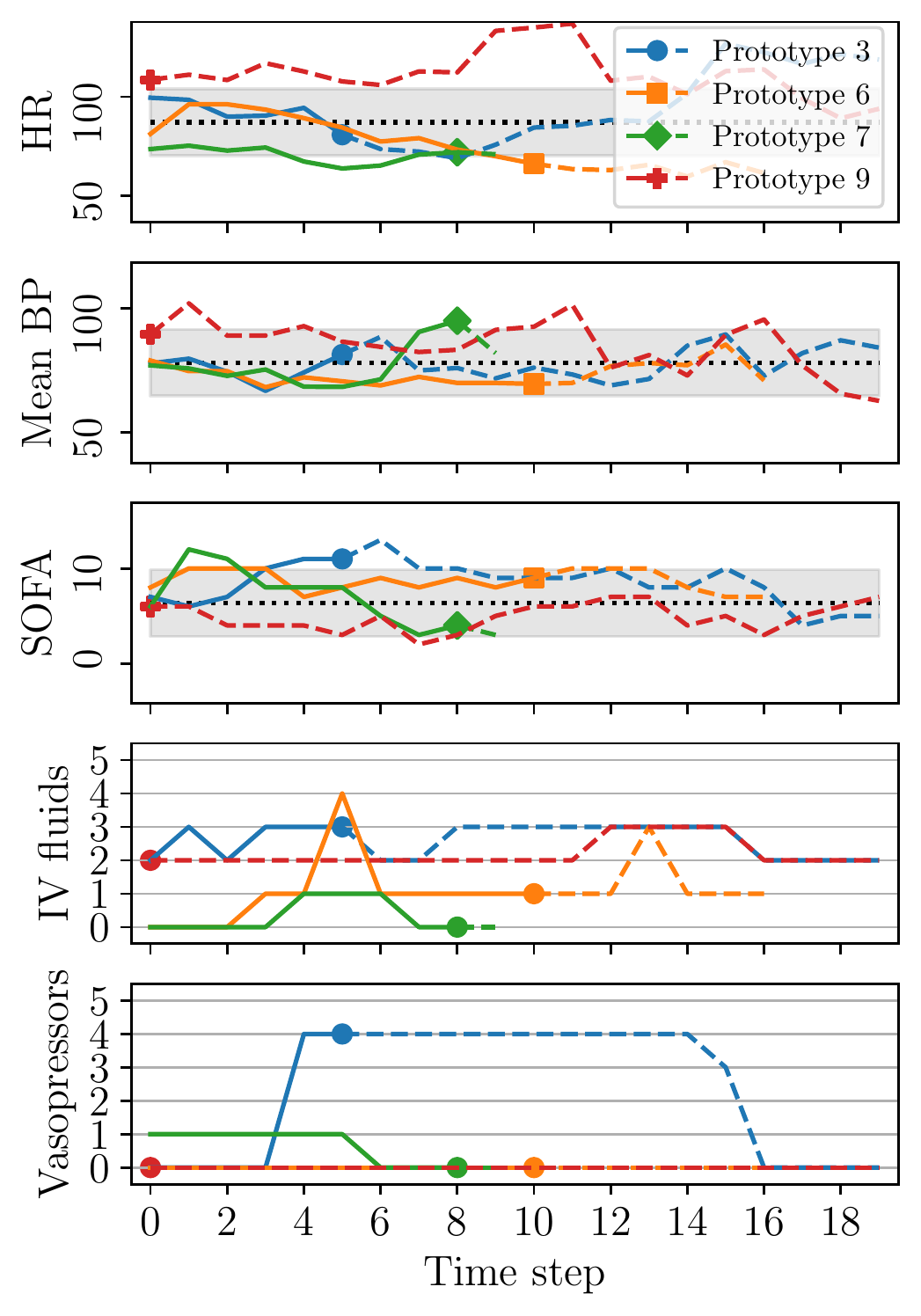}
\caption{Vital signs and SOFA score plotted against time for four different prototype patients. The dashed black lines show each feature's average value in the data and the shaded areas mark $\pm 3$ standard deviations. The lower two panels show the actions taken by the physicians.}
\label{fig:sepsis:features_actions}
\end{figure}

We can interpret the prototypes by studying the trajectories of the corresponding patients. In Figure \ref{fig:sepsis:features_actions}, we plot three key features---heart rate (HR), mean blood pressure (BP) and SOFA score\footnote{SOFA: Sequential Organ Failure Assessment.}---as well as the treatment variables against time. Note that the time index of each prototype is marked with filled marker; for example, prototype 3 is the subsequence ending at time 5 of the corresponding patient history. We see for instance that prototype 7 corresponds to a patient who received low doses of IV fluids and vasopressors throughout, and that the prototype 3 patient were treated more aggressively. 

\begin{figure}[t!]
\centering
\includegraphics[width=0.5\columnwidth]{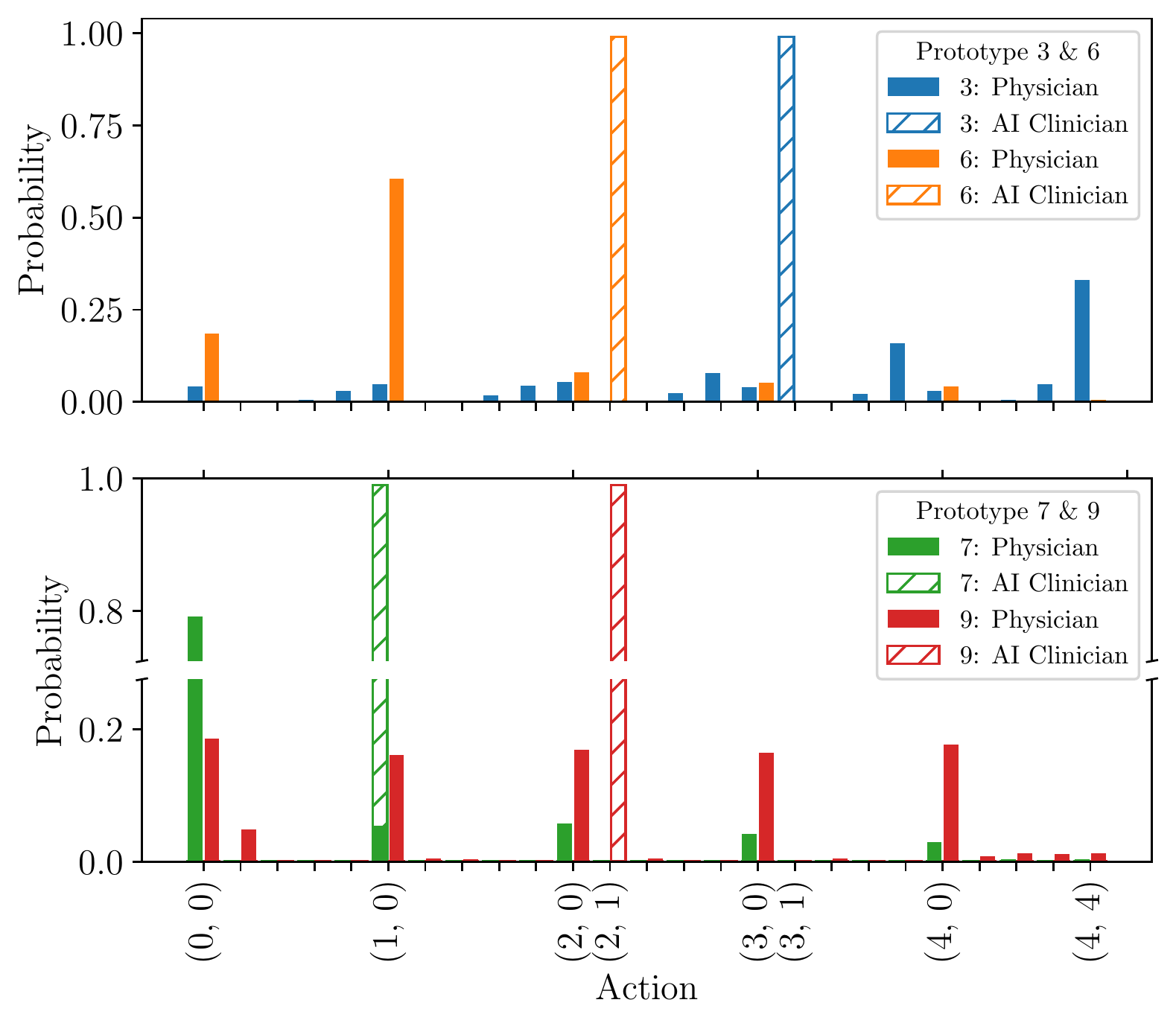}
\caption{The distribution of actions suggested by the physicians (modeled with ProSeNet using $n=10$ and $q=2$) and the AI Clinician for prototypes 3, 6, 7 and 9. Taking prototype 7 as example, the physicians do not usually give any treatment to patients who are similar to this prototype (action (0, 0)), while the AI Clinician suggests giving these patients IV fluids (action (1, 0)).}
\label{fig:sepsis:probas}
\end{figure}

By evaluating the target and behavior policies for each prototype, see \eqref{eq:prototype_policy}, we quickly get an overview of differences between the policies; see Figure \ref{fig:sepsis:probas} where we compare the AI Clinician with the behavior policy for prototypes 3, 6, 7 and 9. Note that the zero-drug policy always suggests action (0, 0) with probability 1. While this action is likely under the behavior policy for prototype 7, it has low probability for prototype 3. From Figure \ref{fig:sepsis:features_actions} we may suspect that the corresponding patient, who received aggressive treatment, likely suffered from severe sepsis. It is reasonable to assume that patients represented by prototype 3 are at greater risk of dying and hence contribute with a negative reward if they are included in the value estimate. For the naive evaluation of the zero-drug policy, however, most of these samples are ignored due to lack of overlap, and the estimated value is probably inflated.

In Figure~\ref{fig:sepsis:probas}, we note a difference between the AI Clinician and the behavior policy for prototype 7, which represents relatively healthy and likely untreated patients. The AI Clinician suggests increasing the dose of fluids for these patients---a rare action under the behavior policy. Such observations are consequently assigned high importance weights. Furthermore, for prototypes 6 and 9, the action (2, 1) of moderate fluids and low vasopressors suggested by the AI Clinician is never observed under the behavior policy---an example of complete lack of overlap. Thus, the estimated value does not reflect this choice of action.

\begin{figure}[t!]
\centering
\includegraphics[width=0.5\columnwidth]{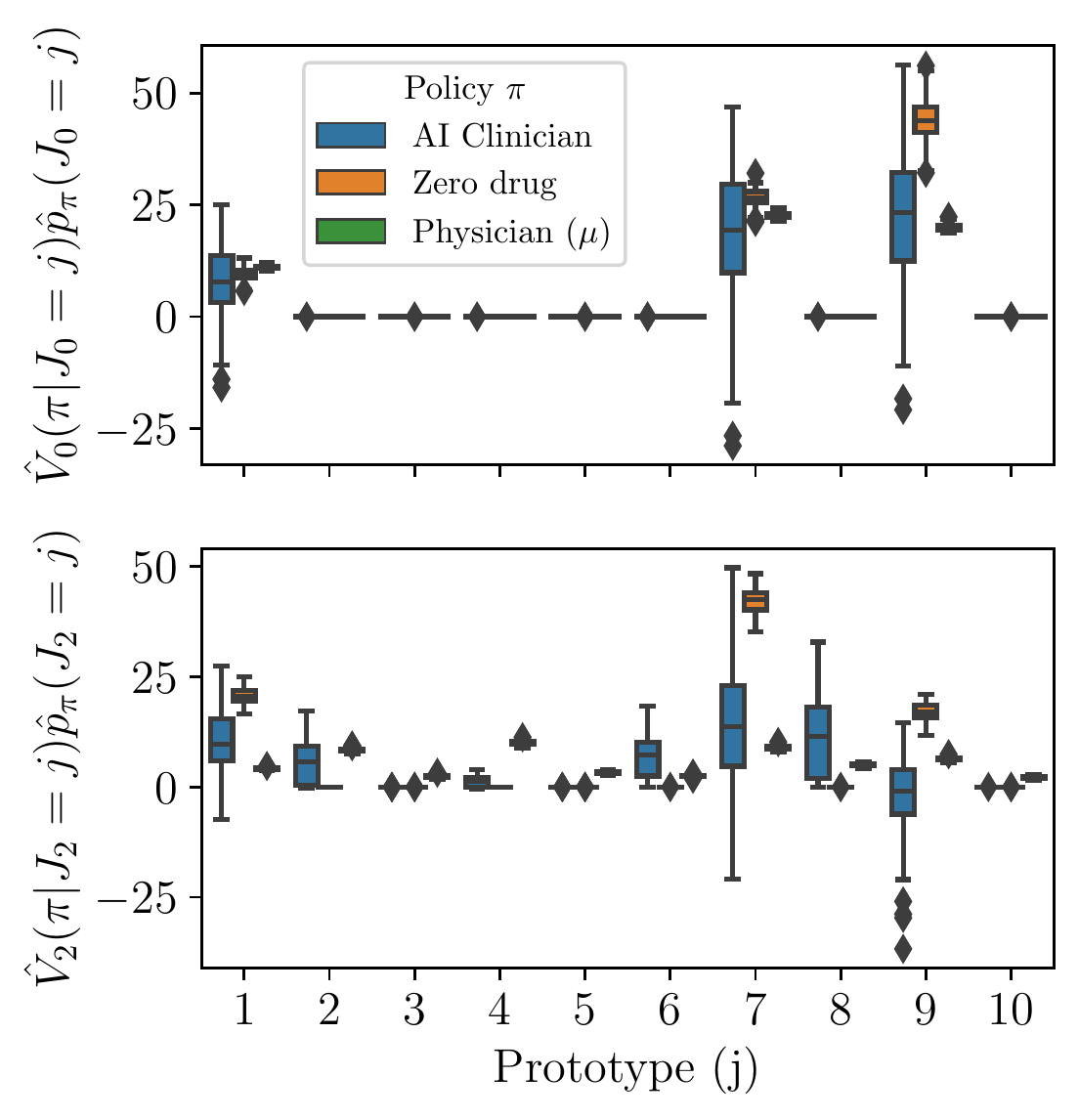}
\caption{Prototype-based contributions to the overall value $V(\pi)$ at time $t=0$ (upper panel) and $t=2$ (lower panel), respectively. We include both target policies and the behavior policy followed by the clinicians (estimated with the ProSeNet model). The boxes are always ordered as in the legend.}
\label{fig:sepsis:VtpJt}
\end{figure}

We have now shown how we can use the prototypes to answer Questions A and B, stated in Section~\ref{sec:is_problems}. It remains to answer Question C: If $\hat{V}(\pi) > \hat{V}(\mu)$, what gives $\pi$ the edge? To do this, we can divide the estimated value into each of the prototypes, as described in Section \ref{sec:protovalues}. In Figure \ref{fig:sepsis:VtpJt}, we estimate $V_{j,t}(\pi)p(J_t=j)$ for each prototype of the ProSeNet model, for $t=0$ and $t=2$. At the initial time step, we see that only prototypes 1, 7 and 9 contribute to the value estimates. This is reasonable since these prototypes correspond to relatively healthy patients, and we expect most patients to be relatively healthy from the beginning. Interestingly, both the AI Clinician and the zero-drug policy have higher value than the behavior policy for prototype 9. It should be stressed that this result only applies to the first time step. Already at $t=2$, we see that the value of the AI Clinician for prototype 9 is lower than that for $\mu$. For this time step, we observe a slight positive effect of following the AI Clinician for prototype 6. However, given the lack of overlap between the AI Clinician and $\mu$, see Figure \ref{fig:sepsis:probas} for this prototype, the data may not support evaluating the AI Clinician for this prototype.

%%%%%%%%%% Model performance %%%%%%%%%%%

\subsection{Analyzing the quality of the estimators}

\begin{figure}
\centering
\begin{minipage}{.48\textwidth}
  \centering
  \includegraphics[width=\linewidth]{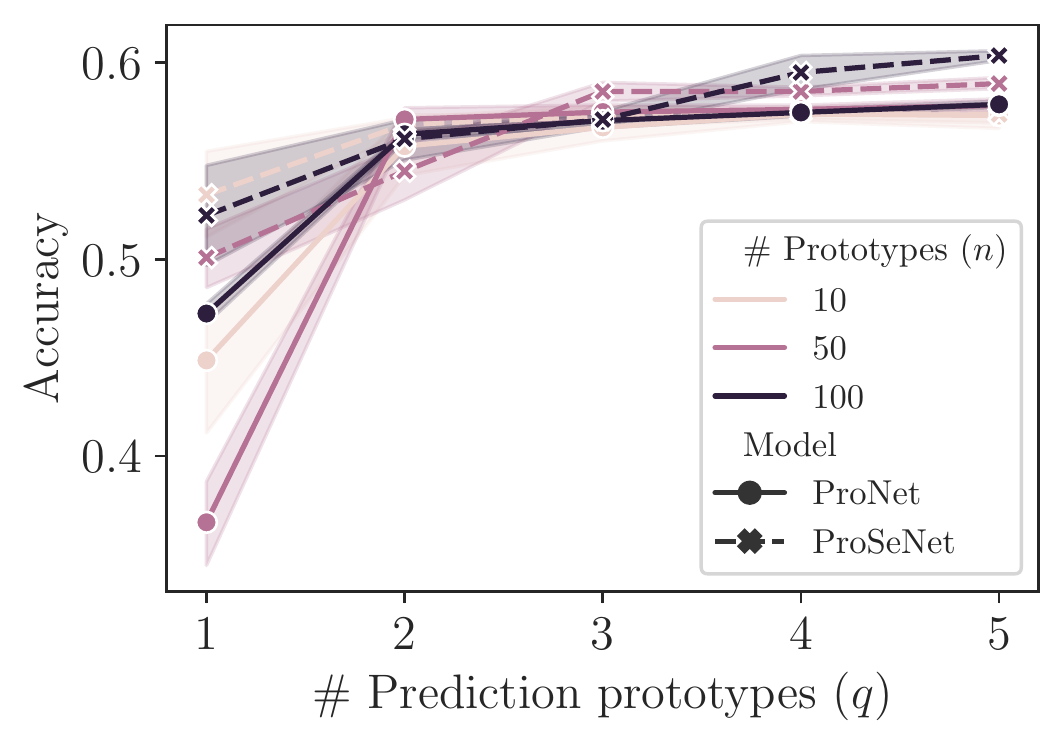}
  \caption{A comparison between ProNet and ProSeNet using a varying number of (prediction) prototypes $(q)$ $n$.}
  \label{fig:sepsis:pronet_prosenet}
\end{minipage}
\hfill
\begin{minipage}{.48\textwidth}
  \centering
  \includegraphics[width=\linewidth]{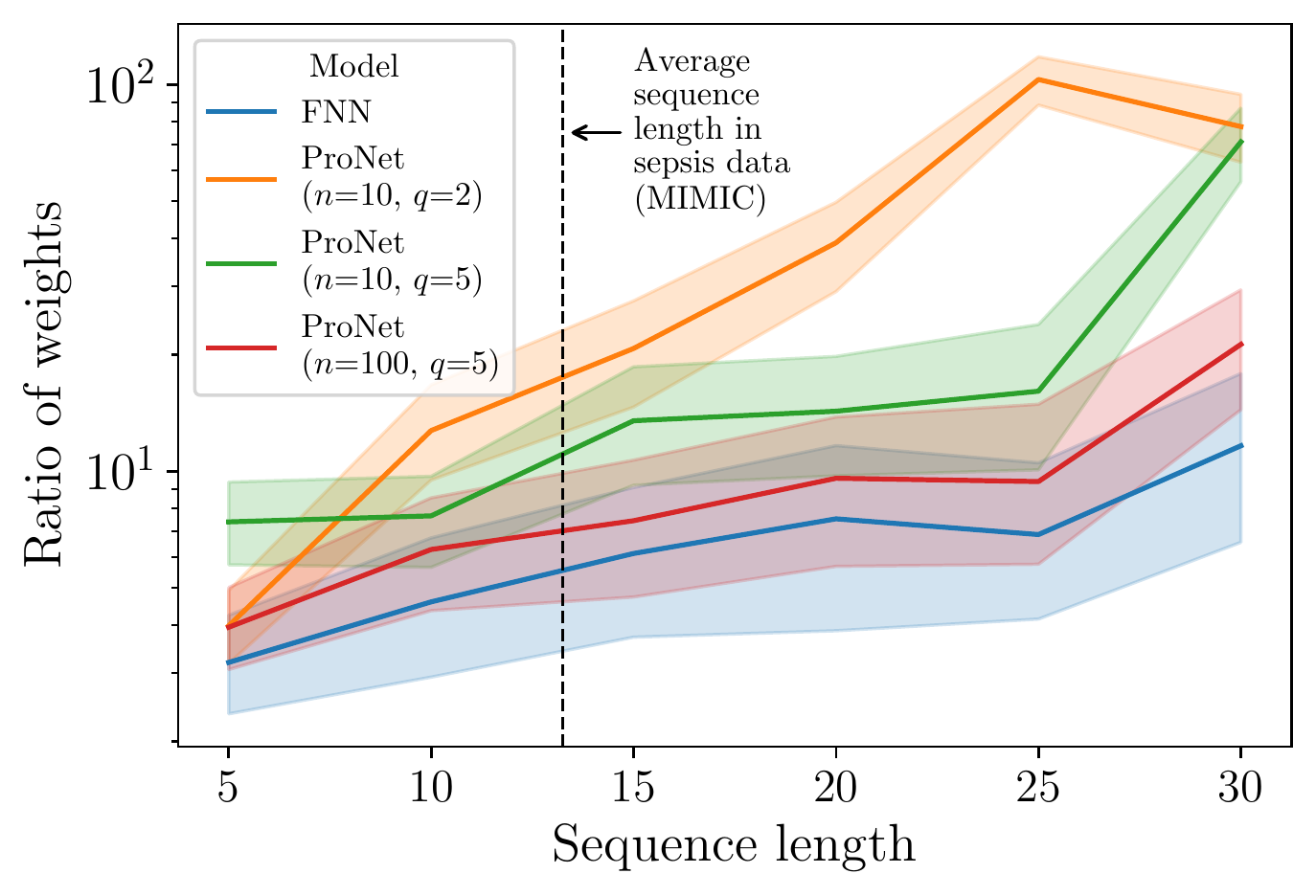}
  \caption{The ratio of the importance weights under $\hat{\mu}$ and $\mu$ for increasing sequence lengths. We compute $\hat{\mu}$ using both a feedforward neural network and ProNet models with different prototype settings.}
  \label{fig:sepsis:bias}
\end{minipage}
\end{figure}

We have argued that we can use prototype learning to obtain a transparent estimate of the behavior policy. We can then inspect the prototypes to reason about how well the data supports evaluation of various target policies. However, while introducing transparency, the use of prototypes imposes restrictions on the model, possibly increasing the approximation error. What can be said about the bias induced by the prototypes? In Figure \ref{fig:sepsis:pronet_prosenet}, we show the accuracy of ProNet and ProSeNet in predicting actions $A_t$ given histories $H_t$---approximating $p_\mu(A\mid H_t)$---on the sepsis test data for a varying number of prototypes $n$ and prediction prototypes $q$. Overall, the sequential model, making use of the entire history $H_t$, performs best, especially for $q=1$ and $q \geq 4$. Interestingly, the effect of increasing the number of prototypes from 10 to 50 or even 100 is rather small. Using only two prediction prototypes works well for this dataset.

In Table \ref{tab:sepsis:performance}, we compare the prototype models with $n=10$ and $q=2$ to the baseline estimators in approximating $p_\mu(A\mid H_t)$. Here, we report accuracy, the area under the ROC curve (AUC) and the static calibration error (SCE)~\citep{nixon2019measuring}, a multiclass extension of the expected calibration error. The prototype models are superior to the (regularized) LR model but they perform slightly worse than black-box models, RF, FNN and RNN. However, as we saw in Figure \ref{fig:sepsis:pronet_prosenet}, with increased number of prototypes, ProSeNet has the capacity to approach the performance of these models, at least in terms of accuracy. See Appendix B for details about, e.g., hyperparameters of the models.

\begin{table*}[t!]
\caption{A summary of test-set performance of different estimators of the behavior policy $p_\mu(A\mid H_t)$. For ProNet and ProSeNet, $n=10$ and $q=2$. The 95 percent confidence intervals are calculated from 1000 bootstraps.}
\centering
\begin{tabular}{@{}llll@{}} \toprule
Model  & Accuracy ($\uparrow$)       & SCE ($\downarrow$)          & AUC ($\uparrow$)           \\ \midrule
LR & 0.38 (0.38, 0.39) & 0.0112 (0.0110, 0.0115) & 0.88 (0.88, 0.88) \\
RF & 0.62 (0.61, 0.62) & 0.0037 (0.0034, 0.0039) & 0.93 (0.93, 0.93) \\
FNN & 0.61 (0.61, 0.61) & 0.0041 (0.0039, 0.0044) & 0.93 (0.92, 0.93) \\
ProNet ($n=10$, $q=2$) & 0.56 (0.55, 0.56) & 0.0069 (0.0067, 0.0072) & 0.90 (0.90, 0.90) \\
RNN & 0.62 (0.62, 0.63) & 0.0056 (0.0053, 0.0058) & 0.94 (0.94, 0.94) \\
ProSeNet ($n=10$, $q=2$) & 0.57 (0.57, 0.58) & 0.0057 (0.0054, 0.0059) & 0.91 (0.91, 0.91) \\
\bottomrule
\end{tabular}
\label{tab:sepsis:performance}
\end{table*}

%%%%%%%%%% Model bias %%%%%%%%%%%

\subsubsection{Bias due to increased sequence length}
\label{sec:simulator}

According to the results presented in the previous section, the use of prototypes introduces a small bias in the estimated propensity. It is natural to ask what this means for the sequential setting, where multiple propensities are multiplied together to form the importance weights. To quantity this effect, we consider the synthetic environment of sepsis management provided by~\cite{oberst2019counterfactual}. By sampling a large amount of data from the environment, we estimate the true parameters of the underlying Markov decision process. We then learn an optimal behavior policy using policy iteration.

We collect trajectories of the behavior policy of various lengths, from 5 to 30 time steps, and for each trajectory length, we estimate the behavior policy from data using an FNN as well as ProNet models with different prototype settings (see Appendix B for details). Given an arbitrary target policy $\pi$, we can now estimate $V(\pi)$ using both the true behavior policy $\mu$ and its estimators $\hat{\mu}$. Any difference in the value estimates stems from the difference in the importance weights. In Figure \ref{fig:sepsis:bias}, we plot the ratio of the weights under $\hat{\mu}$ and $\mu$ against the trajectory length for selected estimators of $\mu$.\footnote{Note that the probabilities under any target policy $\pi$ cancel when considering the ratio of the weights.} As the horizon increases, we see that the ratio for the simplest prototype model with $n=10$ and $q=2$, on average, greatly separates from that for the plain FNN. However, with additional (prediction) prototypes, the effect is smaller.

To estimate the effect on the value estimate, we learn a target policy $\pi$ from trajectories of length 15 (close to the average sequence length in the data from the MIMIC-III database) and compute differences between the true value of $\pi$ and the WIS estimates using the estimators in Figure \ref{fig:sepsis:bias}. Note that we can compute the true value of $\pi$ by simply running it in the simulator. We observe final rewards $r^{i}=\pm 1$ based on survival of the simulated patients. On average, the estimated value has an absolute difference from the true value that amounts to 0.40 for the FNN (standard deviation 0.24), 0.46 (0.26) for ProNet with $n=10$ and $q=2$, 0.52 (0.31) for ProNet with $n=10$ and $q=5$ and 0.44 (0.28) for ProNet with $n=100$ and $q=5$.

%%%%%%%%%%%%%%%%%%%%%%%%%%%%%%%%%%
%%%%%%%%%% Related work %%%%%%%%%%
%%%%%%%%%%%%%%%%%%%%%%%%%%%%%%%%%%

\section{Related work}

% High variance of IS and related issues
Issues with importance sampling methods for OPPE are well known. Several works aim at describing issues related to high variance~\citep{gottesman2019guidelines}, or mitigating them using methodological advances~\citep{precup2000eligibility,thomas2016data,jiang2016doubly,schneeweiss2009high,swaminathan2015counterfactual}. Others move the goalposts, using the weights to identify a new study  population for which the policy's value can be efficiently estimated~\citep{li2018balancing,fogarty2016discrete}. \citet{oberst2020characterization} emphasize the value of interpretability in this endeavour to communicate the generalizability of the estimate. Our method is compatible with each approach, allowing for transparent descriptions of variance issues, identifying new study populations and for use as plug-in estimates.

% Interpretability
Interpretability is a an important component of learning systems deployed in increasingly critical functions~\citep{rudin2019stop,lipton2018mythos}. Rule-based estimators, such as rule list~\citep{wang2015falling} and decision trees, are often favored for their short descriptions but generalize poorly to sequential inputs which are the focus of this work. \citet{gottesman2020interpretable} proposed an approach for interpretable OPPE which highlights transitions in data whose removal would have a large impact on the estimate. This approach is related to ours but answers a different set of questions.  

% Matching
Evaluating policies using direct sample-to-sample comparison has a long tradition in policy evaluation through the use of matching estimators of causal effects, see e.g.,~\citep{rosenbaum1983central,rubin2006matched,kallus2020generalized}. While favored for its transparency, this approach is typically only used to compare two deterministic policies such as ''treat all'' or ''treat none''. Matching often relies either on specifying a similarity function in advance or on an estimate of the behavior policy. In high-dimensional settings, this often leads to bias or lost interpretability. Our approach aims to combine the transparency of matching estimators with the flexibility of representation learning methods.

%%%%%%%%%%%%%%%%%%%%%%%%%%%%%%%%
%%%%%%%%%% Conclusion %%%%%%%%%%
%%%%%%%%%%%%%%%%%%%%%%%%%%%%%%%%

\section{Conclusion}

We have studied off-policy policy evaluation (OPPE) using importance sampling (IS) in the case where the behavior policy $\mu$ is unknown and must be estimated from data. We motivated why IS can be difficult and identified three questions about the IS estimate that are likely to remain unanswered when the behavior policy is estimated using black-box models: Which observations contribute to the IS estimate? In which situations is overlap violated? If $\hat{V}(\pi)>\hat{V}(\mu)$, what gives the target policy $\pi$ the edge? We proposed performing OPPE using prototype learning to better answer these questions and we illustrated our method for the management of sepsis. To quantify the bias induced by the prototypes, we used a simulated environment of sepsis management. In future work, we hope to provide theoretical results for the approximation error. In particular, how does the error from a prototype-based estimate of the behavior policy propagate to the estimated policy value?

%%%%%%%%%%%%%%%%%%%%%%%%%%%%%%%%%%
%%%%%%%%%% Bibliography %%%%%%%%%%
%%%%%%%%%%%%%%%%%%%%%%%%%%%%%%%%%%

\bibliography{main}

%%%%%%%%%%%%%%%%%%%%%%%%%%%%%%
%%%%%%%%%% Appendix %%%%%%%%%%
%%%%%%%%%%%%%%%%%%%%%%%%%%%%%%

\clearpage
\appendix

\section{The prototype model}
\label{app:reg}

In Section \ref{sec:protomodel}, we only briefly described the objective function
\begin{equation*}
    J(\Theta) = \mathrm{NLL}(\cD; \Theta) + \lambda_{d}R_{d}(\Theta) + \lambda_{c}R_{c}(\Theta) + \lambda_{e}R_{e}(\Theta),
\end{equation*}
where $\mathrm{NLL}(\cD; \Theta)$ is the negative log-likelihood, $R_{d}(\Theta)$, $R_{c}(\Theta)$ and $R_{e}(\Theta)$ are regularization terms and $\lambda_{d}$, $\lambda_{c}$ and $\lambda_{e}$ are regularization parameters. $\Theta$ denotes the set of model parameters, i.e., the parameters of the encoding network $e$, the weights $B, c$ and the prototypes $\tH$.

For a given dataset $\cD = ((h^1_{t_1}, a^1_{t_1}), \ldots, (h^m_{t_m}, a^m_{t_m}))$, drawn according to a distribution $p_{\mu}$, the NLL loss of the estimate $\hat{p}_{\mu}$, parameterized in $\Theta$, is defined as
\begin{equation*}
    \mathrm{NLL}(\cD; \Theta) = -\frac{1}{m} \sum_{i=1}^{m}
    \log{\left(\hat{p}_{\mu}(A_{t}=a_{t_{i}}^{i} \mid H_{t}=h_{t_{i}}^{i})\right)}.
\end{equation*}
Furthermore, the regularization terms are defined as follows:
\begin{itemize}
    \item The \textbf{diversity} regularization $R_{d}(\Theta) = \sum_{i=1}^{n}\sum_{j=i+1}^{n}\text{max}\left(0, d_{\mathrm{min}}-d(\tz_{i}, \tz_{j})\right)^{2}$, where $d(z, z^{\prime}) = \|z-z^{\prime}\|_{2}$, penalizes latent prototypes that are too close to each other. The parameter $d_{\mathrm{min}}$ is a tunable hyperparameter in our experiments.
    \item The \textbf{clustering} regularization $R_{c}(\Theta) = \sum_{h\in \cD} \underset{i}{\text{min}}\,d(\tz_{i}, e(h))^2$ encourages the encoded histories to approach the most similar latent prototypes, which creates a clustering structure in the latent space.
    \item The \textbf{evidence} regularization $R_{e}(\Theta) = \sum_{i=1}^{n}\underset{h\in \cD}{\text{min}}\:d(\tz_{i}, e(h))^{2}$ encourages the latent prototypes to approach the encodings that are most similar.
\end{itemize}

\subsection{Prototype value}
\label{app:protvalue}

Below, we derive a statistical estimator for the value of policy $\pi$ for prototype $j$ at time $t$ using observations under $\mu$. First, we have
\begin{align*}
    V_{j,t}(\pi) & \coloneqq  \E_{\pi} \bigg[\sum_{t' \geq t} R_{t'} \Bigm\vert J_t = j \bigg] \\
    & = \E_{\pi} \bigg[ \frac{p(J_t = j \mid H_t)}{p_\pi(J_t=j)}  \sum_{t' \geq t} R_{t'} \bigg].
\end{align*}
The equation follows from the fact that $J_t$ is conditionally independent of all other variables given $H_t$. Now, with $W$ importance weights for $\pi$ and $\mu$, 
\begin{equation*}
    V_{j,t}(\pi) = \E_{\mu} \bigg[ \frac{p(J_t = j \mid H_t)}{p_\pi(J_t=j)} W \sum_{t' \geq t} R_{t'}  \bigg].
\end{equation*}
Following standard definitions, 
\begin{equation*}
    p_\pi(J_t=j) = \E_\pi[p(J_t \mid H_t)],
\end{equation*}
which may be identified using importance sampling, 
\begin{equation*}
    p_\pi(J_t ) = \E_\pi[p(J_t \mid H_t)] = \E_\mu[p(J_t \mid H_t)W_t],
\end{equation*}
with $W_t = \prod_{t^{\prime}=0}^t \frac{p_\pi(A_{t'} \mid H_{t'})}{p_\mu(A_{t'} \mid H_{t'})}$. Hence, we may estimate
\begin{equation*}
\hat{p}_\pi(J_t=j) = \frac{1}{m}\sum_{i=1}^m  p(J_t=j \mid H_t = h_{t}^{i})w_{t}^{i}.
\end{equation*}
For trajectories $i$ which end before $t$, we let $\hat{p}(J_t=j \mid H_t = h_{t}^{i}) = 0$. 

\section{Experimental details}
\label{app:exp_details}

The prototype model was evaluated on real-world sepsis data extracted from the MIMIC-III database \citep{mimiciii} as well as synthetic data generated by the simulator provided by \cite{oberst2019counterfactual}. Here, we give details about the experiments. To produce the results presented in this paper, we needed about 750~core-hours of computational time. The neural networks were implemented in PyTorch and trained on GPU (Nvidia Tesla T4). Other models were implemented using scikit-learn.

\subsection{MIMIC-III}

We extracted the dataset of patients suffering from sepsis from the MIMIC-III database using the code provided by \cite{komorowski2018artificial}.\footnote{The code is available at \url{https://github.com/matthieukomorowski/AI_Clinician}.} This dataset contains the features listed in Supplementary Table 2 in \cite{komorowski2018artificial} as well as the total fluid intake and the total urine output. We learned the target policy, the so-called AI Clinician, using the Matlab code provided by \cite{komorowski2018artificial}. To evaluate the 500 candidate policies, we used only the MIMIC test data and not data from the eICU Research Institute Database. We used the train-test split associated with the best performing candidate in our experiments.

We trained the estimators of the behavior policy using a subset of the available features: heart rate, systolic blood pressure, diastolic blood pressure, mean blood pressure, shock index, hemoglobin, BUN, creatine, urine output over 4 hours, pH, base excess, bicarbonate, lactate, $\text{PaO}_{\text{2}}/\text{FiO}_{\text{2}}$ ratio, age, Elixhauser index and SOFA score. In addition, we included the treatment doses (vasopressors and IV fluids) over the previous 4 hours. At the first time step, these values were set to 0.
 
\subsubsection{Model parameters}

The neural networks were fitted to the training data using similar architectures. For ProNet, we used an FNN-based encoder with two layers (each of size 64) and ReLu as activation function. For ProSeNet, we used an RNN-based encoder with two layers (each of size 64) and tanh as activation function. The FNN baseline had two layers (each of size 64) and ReLu as activation function. The RNN baseline had 2 layers (each of size 64) and tanh as activation function. A linear layer was added to the baseline models to obtain predictions of the right shape. 

We trained all neural networks over 400 epochs, using a batch size of 64 for RNN and ProSeNet, and 1024 for FNN and ProNet. For optimization, the Adam algorithm was used with default parameters, learning rate 0.001 and weight decay 0.001. The NLL was used as loss function. For ProNet and ProSeNet, we selected parameters of the diversity regularization $(d_{\textrm{min}}, \lambda_{d})$ by performing 3-fold cross-validation over a grid of points in the parameter space $\{1, 2, 3, 4, 5\}\times \{0.00001, 0.0001, 0.001, 0.01, 0.1\}$. Note that these parameters were optimized for each combination of prototypes $n$ and prediction prototypes $q$ in our experiments. The parameters $\lambda_{c}$ and $\lambda_{e}$ were set to 0.001, and we performed the projection step, see \eqref{eq:projection}, every fifth epoch.

For LR and RF, we searched for optimal models using 3-fold cross-validation, considering the following parameter values:
\begin{itemize}
    \item LR: regularization: $\{\mathrm{L1}, \mathrm{L2}\}$; regularization strength: 10 numbers spaced evenly on a log-scale from \SI{1e-4} to \SI{1e4};
    \item RF: maximum tree depth: $\{5, 10, 15, 20, \mathrm{None}\}$.
\end{itemize}

All models were calibrated using sigmoid calibration on a held-out validation set (\SI{25}{\percent} of the training data).

\subsection{Sepsis simulator}

To estimate the bias induced by the prototypes, we used the sepsis simulator provided by \cite{oberst2019counterfactual}.\footnote{The simulator is publicly available at \url{https://github.com/clinicalml/gumbel-max-scm/tree/sim-v2}.} We used the full state representation consisting of five discrete variables (a binary diabetes indicator and four ordinal-valued vital signs (heart rate, systolic blood pressure, blood glucose level and blood oxygen level)) as well as the previously administered treatments. There is a probability of 0.2 that a randomly initialized patient has diabetes. Furthermore, there are three binary treatment variables---antibiotics, vasopressors and ventilation---resulting in an action space of size eight. A simulated patient is discharged if the patient has normal vitals and is not given any treatments; discharge results in a positive reward. Death is associated with a negative reward and occurs if at least three vitals are abnormal. In all other cases, the reward is zero. We refer to the code for details about for example the levels of the vitals and the transition probabilities of the Markov decision process (MDP). 

For the experiment presented in Figure \ref{fig:sepsis:bias}, we used the notebook \verb|learn_mdp_parameters| provided by \cite{oberst2019counterfactual} to estimate the true parameters of the MPD. We then learned an optimal behavior policy using policy iteration. The policy was softened so that all state-action pairs had a nonzero probability of being chosen. We generated trajectories of varying length from this policy to estimate the effect of bias. Some trajectories ended prematurely due to discharge or death and we therefore ensured that the number of state-action pairs were roughly the same for all sequence lengths. Specifically, for each sequence length, we generated \SI{20000} state-action pairs for training, sigmoid calibration and evaluation, respectively. We trained ten replications of the models using different collections of samples.

The target policy referred to in Section \ref{sec:simulator} was learned by first estimating the MDP parameters from the generated training samples and then solving the estimated MDP using policy iteration. The policy was softened to avoid zero probabilities.

\subsubsection{Model parameters}

Since we used the full state representation, we could safely make the Markov assumption and estimate the behavior policy using FNN-based models. For ProNet, we used an FNN-based encoder with two layers (each of size 64) and ReLu as activation function. The FNN baseline had two layers (each of size 64) and ReLu as activation function. A linear layer was added to obtain predictions of the right shape. We trained all models over 30 epochs, using a batch size of 128. For optimization, the Adam algorithm was used with default parameters, learning rate 0.001 and weight decay 0.001. Again, we selected parameters of the diversity regularization $(d_{\textrm{min}}, \lambda_{d})$ by performing 3-fold cross-validation over a grid of points in the parameter space $\{1, 2, 3, 4, 5\}\times \{0.00001, 0.0001, 0.001, 0.01, 0.1\}$. The parameters $\lambda_{c}$ and $\lambda_{e}$ were set to 0.001, and we performed the projection step every fifth epoch.

\end{document}

%% file: preamble.tex
\usepackage[T1]{fontenc}   
\usepackage[utf8]{inputenc}   
\usepackage[english]{babel}
\usepackage[margin=1.5in]{geometry}
\usepackage{authblk}

\usepackage[round]{natbib}
\usepackage{amssymb}
\usepackage{mathtools}
\usepackage{siunitx}

\usepackage{graphicx}
\usepackage{float}
\usepackage{booktabs}

\usepackage{hyperref}
\usepackage{enumerate}

\DeclareMathOperator*{\argmin}{arg\,min}

\def\IS{\mathrm{IS}}
\def\WIS{\mathrm{WIS}}

\def\cA{\mathcal{A}}
\def\cD{\mathcal{D}}
\def\cH{\mathcal{H}}
\def\cX{\mathcal{X}}
\def\cZ{\mathcal{Z}}

\def\th{\tilde{h}}
\def\tH{\tilde{H}}
\def\tz{\tilde{z}}
\def\tZ{\tilde{Z}}

\def\E{\mathbb{E}}
\def\bbR{\mathbb{R}}

\def\hp{\hat{p}}